%% file: paper-draft.tex
\documentclass[sigconf]{acmart}

\input{math_commands.tex}

\usepackage{url}
\usepackage{graphicx}
\usepackage{color}
\usepackage{multirow}
\usepackage{subfigure}
\usepackage{ulem}
\usepackage{enumitem}
\usepackage{makecell}
\usepackage{tabularx}
\usepackage{threeparttable}
\usepackage{arydshln}
\usepackage{soul}
\usepackage{color, xcolor}
\definecolor{mylightgreen}{RGB}{200,255,200}

\DeclareMathOperator*{\mlp}{MLP}
\DeclareMathOperator*{\transformerenc}{Transformer-Enc}
\DeclareMathOperator*{\pool}{Pool}

\DeclareMathOperator*{\kl}{KL}

\AtBeginDocument{%
  \providecommand\BibTeX{{%
    \normalfont B\kern-0.5em{\scshape i\kern-0.25em b}\kern-0.8em\TeX}}}
    
\copyrightyear{2024}
\acmYear{2024}
\setcopyright{acmlicensed}\acmConference[WWW '24]{Proceedings of the ACM Web Conference 2024}{May 13--17, 2024}{Singapore, Singapore}
\acmBooktitle{Proceedings of the ACM Web Conference 2024 (WWW '24), May 13--17, 2024, Singapore, Singapore}
\acmDOI{10.1145/3589334.3645471}
\acmISBN{979-8-4007-0171-9/24/05}

\begin{document}

\title{Explainable Fake News Detection With Large Language Model via Defense Among Competing Wisdom}

\author{Bo Wang}
\affiliation{%
  \institution{School of Artificial Intelligence, \\Jilin University}
  \city{Changchun}
  \country{China}
  }
\affiliation{%
  \institution{Hong Kong Baptist University}
  \city{Hong Kong SAR}
  \country{China}
  }
\email{wangbo21@mails.jlu.edu.cn}

\author{Jing Ma$^*$}
\author{Hongzhan Lin}
\affiliation{%
  \institution{Hong Kong Baptist University}
  \city{Hong Kong SAR}
  \country{China}
  }
\email{majing@comp.hkbu.edu.hk}
\email{cshzlin@comp.hkbu.edu.hk}

\author{Zhiwei Yang}
\affiliation{%
  \institution{Guangdong Institute of Smart Education, \\Jinan University}
  \city{Guangzhou}
  \country{China}
  }
\email{yangzw18@mails.jlu.edu.cn}

\author{Ruichao Yang}
\affiliation{%
  \institution{Hong Kong Baptist University}
  \city{Hong Kong SAR}
  \country{China}
  }
\email{csrcyang@comp.hkbu.edu.hk}

\author{Yuan Tian$^*$}
\affiliation{%
  \institution{School of Artificial Intelligence, \\Jilin University}
  \city{Changchun}
  \country{China}
  }
\email{yuantian@jlu.edu.cn}

\author{Yi Chang$^*$}
\thanks{*Joint Corresponding Author}
\affiliation{%
  \institution{School of Artificial Intelligence, \\Jilin University}
  \city{Changchun}
  \country{China}
  }
\email{yichang@jlu.edu.cn}

\renewcommand{\authors}{Bo Wang, Jing Ma, Hongzhan Lin, Zhiwei Yang, Ruichao Yang, Yuan Tian, Yi Chang}
\renewcommand{\shortauthors}{Bo Wang et al.}

\begin{abstract}

Most fake news detection methods learn latent feature representations based on neural networks, which makes them black boxes to classify a piece of news without giving any justification. Existing explainable systems generate veracity justifications from investigative journalism, which suffer from debunking delayed and low efficiency. Recent studies simply assume that the justification is equivalent to the majority opinions expressed in the wisdom of crowds. However, the opinions typically contain some inaccurate or biased information since the wisdom of crowds is uncensored.
To detect fake news from a sea of diverse, crowded and even competing narratives, in this paper, we propose a novel defense-based explainable fake news detection framework. Specifically, we first propose an evidence extraction module to split the wisdom of crowds into two competing parties and respectively detect salient evidences. To gain concise insights from evidences, we then design a prompt-based module that utilizes a large language model to generate justifications by inferring reasons towards two possible veracities. Finally, we propose a defense-based inference module to determine veracity via modeling the defense among these justifications. Extensive experiments conducted on two real-world benchmarks demonstrate that our proposed method outperforms state-of-the-art baselines in terms of fake news detection and provides high-quality justifications.
\end{abstract}



\maketitle

\section{Introduction} \label{sec:intro}

\begin{figure}[t]
    \centering
    \includegraphics[width=0.42\textwidth]{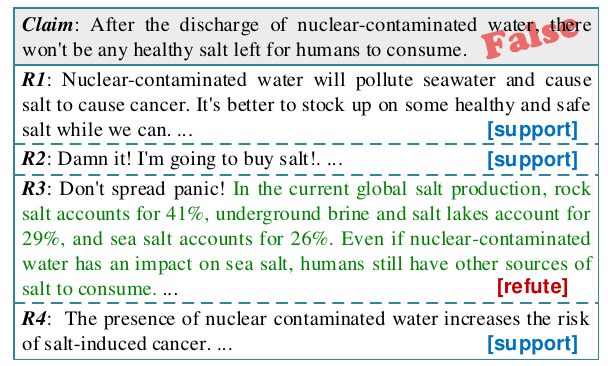}
    \vspace{0mm}
	\caption{\small A false claim from the Sina Weibo. The comparison of informativeness and soundness between two competing parties serves as an indicator of veracity. 
	}
	\label{fig:intro_illustration} 
\end{figure}

The proliferation of fake news on social media has become a remarkable concern, leading to detrimental effects on individuals and society. For example, during the global COVID-19 pandemic, a piece of spurious news claiming that ``\textit{the COVID-19 vaccine can induce serious side effects and potentially result in death}\footnote{\url{ https://www.bbc.com/news/53525002}}'' attracted the public's attention, leading to people's hesitancy and resistance towards vaccine uptake, and thus seriously impacted the virus containment effort and overwhelmed healthcare systems around the world. 
Fortunately, as the truth was consistently justified by the official media and investigative journalism, the public recognized the claim as fake. 
This indicates the positive role of solid justification in restricting the social harmfulness caused by fake news. 
However, relying solely on investigative journalism to enable the public to detect fake news is not a realistic approach. Such a labor-intensive and time-consuming manner limits the coverage and is debunking delayed. 
Thus, it is urgent to develop automated methods to detect fake news and provide clear justifications timely.

Most previous works for detecting fake news focus on incorporating various information to learn the latent features of fake news via deep networks, e.g., credibility \cite{popat2017truth}, stances \cite{ma_WWW18},  propagation patterns \cite{AA-HGNN_Ren_ICDM20}, extra knowledge \cite{dun2021kan}, out-of-domain data \cite{hongzhan_naacl22}, etc. Despite their success in detecting, limited by their black-box nature, they are unable to provide any justification, which is essential to the public. 
To address this problem, some studies are dedicated to explainable fake news detection (EFND) that aims to generate veracity prediction and explanations at the same time \cite{EFKD_survey_KotonyaT_coling20}. 
Many works provide their interpretability by highlighting salient words, phrases, or sentences in relevant reports via attention mechanism \cite{DeClarE_EMNLP18, conflict_wu_aaai21,Nie_AAAI19,dEFEND_KDD19}. 
However, these works only uncover regions with high contributions for the final prediction, lacking intuitive and credible explanations. 
As human justification brings great improvement in veracity prediction \cite{LIAR-PLUS_Alhindi_18}, some works use debunked reports to generate explanations. \citet{GenFE_ACL20} obtains explanations by summarizing from debunked reports, but suffers from debunking delayed and low efficiency. To mitigate this problem, motivated by the effectiveness of the wisdom of crowds in fact-checking \cite{crowds_wisdom-Allen_2021}, \citet{zhiwei22coling} assumes the majority of opinions expressed in the raw reports are equivalent to the justifications and extracts explanations from them. However, unverified raw reports typically contain inaccurate or biased information since the wisdom of crowds is uncensored. The ill-considered assumption leads to misleading results that are biased towards the majority opinion in raw reports. 
Therefore, it remains a challenge to effectively leverage the rich wisdom expressed in raw reports to support EFND.

Recent detection in the field of stance detection implicitly suggests that the different insights in various raw reports are crucial signals in the quest for truth \cite{jin_AAAI16, ma_WWW18,ruichao_sigir22}. Inspired by it, we propose to split the wisdom into two distinct parties, which allows the detection to rely on the quality of wisdom rather than its quantity. 
Take a concrete example, as shown in Figure~\ref{fig:intro_illustration}, there are two competing parties to the claim. For the supporting party, R1 and R4 both briefly discuss the risk of salt leading to cancer, and R2 echoes the claim without additional information. In contrast, R3 provides the refuting party with detailed evidence to illustrate its unique viewpoint, which is solid and persuasive. 
Based on the observation, we assume that the reports indicating truthfulness could exhibit higher quality of informativeness and soundness compared to those conveying inaccurate information. As a result, the veracity of news can be ascertained through a comparative analysis between two competing parties. 
Therefore, how to effectively split the wisdom into two parties from raw reports and then capture their quality divergence, is a critical problem for enhancing EFND.

To deal with the above issues, we propose a defense-based explainable fake news detection framework, which strives to capture the divergence between the competing wisdom reflected in raw reports and pursue the veracity of claims in a defense-like way. 
Specifically, we first propose an evidence extraction module to split the wisdom of crowds into two competing parties, from which we detect salient evidences, respectively. 
Since the wisdom of each competing party is massive, it is formidable to identify the divergence by directly using the competing evidences, thereby raising the demand for streamlined summarization. 
Inspired by the dominating performance of large language models (LLMs) \cite{chatgpt,llama,llama2,GPT-4}, based on the respective evidences, we then design a prompt-based module to generate justifications by inferring reasons towards two possible veracities. Benefiting from the powerful reasoning and generating capacity of LLMs, we obtain the summarized wisdom of both parties in natural language, allowing for explicit comparisons. 
Finally, to capture the winner in quality comparisons, namely the party indicating the veracity of the claim, we propose a defense-based inference module to determine veracity by modeling the defense among these justifications. 
In this manner, the final justification for the verdict is adaptively selected from these justifications. Our main contributions are summarized as follows: 
\begin{itemize}[leftmargin=10pt]
    \vspace{0mm}
    \item We develop a defense-based framework to utilize the rich competing wisdom naturally contained in raw reports, mitigating the majority bias problem from which existing works suffer. 
    \vspace{0mm}
    \item By integrating the powerful reasoning capabilities of LLMs, our model can derive explanations comparable to those of human experts without any debunked reports' supervision. 
    \vspace{0mm}
    \item We achieve new state-of-the-art results on two real-world explainable fake news detection datasets and demonstrate the quality of the explanations with extensive analyses. 
\end{itemize}

\begin{figure*}[t]
    \centering
    \includegraphics[width=0.9\textwidth]{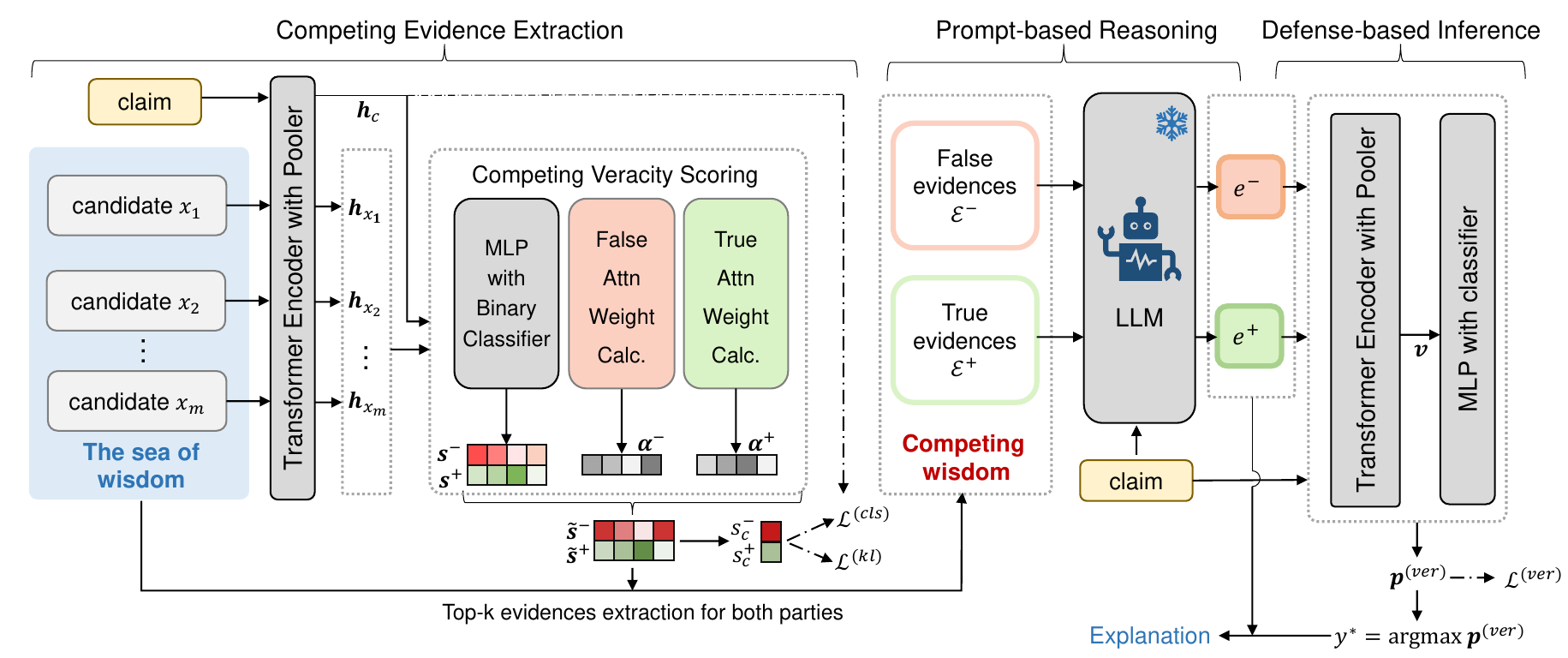}
    \vspace{0mm}
	\caption{\small An overview of the proposed LLM-equipped defense-based explainable fake news detection (L-Defense) framework.
	}
	\label{fig:model_architecture} 
\end{figure*}
\section{Related Work} \label{sec:related_work}
\subsection{Explainable Fake News Detection}
Much effort has been devoted to investigating the field of EFND in previous studies. 
To bring about some explainability, some works explore attention mechanisms to highlight salient phrases \cite{DeClarE_EMNLP18,conflict_wu_aaai21}, news attributes \cite{XFake_yang_www19}, and suspicious users \cite{GCAN_Lu_and_Li_ACL20}. 
In order to gain more human-readable explanations, some works capture the salient sentences as explanations via sentence-level attention \cite{Nie_AAAI19,HAN_ma_acl19,dEFEND_KDD19}. 
However, these methods merely uncover regions with high contributions to the final veracity prediction rather than view the explanation generation as a dependent task. 
To address it, \citet{EXTABS_EMNLP20_Kotonya} regards the explanation generation task as a pre-trained extractive-abstractive summarization task. 
\citet{GenFE_ACL20} formats the EFND task as a multi-task learning problem, and then trains a joint model based on external debunked reports collected from fact-checking websites. However, the debunking of claims is a labor-intensive and time-consuming process. Such a heavy reliance on debunked reports results in coverage limitations and debunked delays. 
To alleviate this problem, \citet{zhiwei22coling} merely employs the debunked report as a supervised signal in training, and concentrates on the majority opinions from crowds expressed in relevant raw reports to aid in prediction and extract evidence. However, it ignores the inaccurate and biased information in unverified reports, causing a misleading result. 
Therefore, by splitting the wisdom of crowds into two competing parties and introducing a defense-like strategy, we capture and leverage the divergence between the competing wisdom to reach a verdict on the claim.

\subsection{Large Language Model in Fake News Detection} \label{sec:related_LLM}
Recently, LLMs have been proven excellent ability in multiple classification and reasoning tasks \cite{InstructGPT_ouyang_NIPS22,chatgpt,llama2,GPT-4,survey_LLM_eval}. 
Unfortunately, the huge training cost prevents LLMs from keeping up with the latest information, which restricts the application of LLMs in the field of fake news detection with high real-time requirements. 
Several studies experimentally demonstrate there is a gap between LLMs and fine-tuned small language models (SLMs) like BERT \cite{bert_devlin_naacl19}, but also indicate that LLMs hold great potential for detecting fake news \cite{sekf-checker_Li_arxiv23,LLM_SLM_distill_Hu_arxiv23,LLM_biased_su_arxiv23}. 
\citet{sekf-checker_Li_arxiv23} propose a step-by-step framework consisting of a set of plug-and-play modules to facilitate fact news detection. It achieves promising results in a zero-shot setting by purely prompting LLMs with external retrieved knowledge. 
\citet{LLM_SLM_distill_Hu_arxiv23} find that LLMs are suboptimal at veracity judgment but good at analyzing contents. It thus trains the small language model to adaptively acquire insights from LLM-generated rationales in a distillation framework. 
\citet{factllama} use LoRA tuning \cite{lora_ICLR22} to train a LLaMA-based \cite{llama} detector with external retrieved knowledge. 
However, these methods mainly concentrate on the detection of fake news and lack the ability to provide an easily comprehensible explanation to the publics. Also, the methods incorporating external retrieved knowledge still suffer from the majority bias problem. 
In contrast, we employ LLMs to generate justifications towards two possible veracities based on respective evidences. The derived competing justifications are used to detect veracity via a defense-like process.

\section{Proposed Approach} \label{sec:approach}

This section begins with a task definition of explainable fake news detection (EFND). 
Then, we present our \textbf{L}LM-equipped \textbf{D}efense-based \textbf{E}xplainable \textbf{F}ak\textbf{e} \textbf{N}ew\textbf{s} D\textbf{e}tection (L-Defense) framework (as in Figure~\ref{fig:model_architecture}) consisting of three components: a competing evidence extractor (\S \ref{subsec:extractor}), a prompt-based reasoning module (\S \ref{subsec:generator}), and a defense-based inference module (\S \ref{subsec:predictor}). Finally, we discuss the generalization of the proposed L-Defense (\S \ref{subsec:discuss_gene}). 

\paragraph{Task Definition. }
Given a news claim $c$ associated with a veracity label $y$, and its relevant raw reports $\gD = \{d_i\}^{|\gD|}_{i=1}$, where each $d_i = (s_{i,1}, s_{i,2}, \dots, s_{i,|d_i|)}$ denotes a relevant report consisting of a sequence of sentences, EFND aims to predict a veracity label $y^*$ of claim $c$ and provide an explanation $e$ regarding the prediction. For different datasets, $y^*$ can be either in the set \{false, half, true\} or \{pants-fire, false, barely-true, half, mostly-true, true\}.

\subsection{Competing Evidence Extraction} \label{subsec:extractor}
In this subsection, we aim to split the sea of wisdom into two competing parties, and extract respective evidences for subsequent comparison. 
As the only available supervised data is the veracity label of the claim, we conduct a temporary veracity prediction to detect salient evidences for both parties via a veracity scoring module. 
Each claim is assigned a temporary veracity label in \{false, half, true\} based on its real veracity, where ``half'' indicates that the claim contains both true and false aspects.

\subsubsection{Claim and Evidence Representation}
Since a report may contain evidences indicating different veracities, we disaggregate the reports into individual sentences, resulting in a corpus of candidate evidence sentences $\gX = \{x_j\}_{j=1}^m$, where $m = \sum_{d_i \in \gD} |d_i|$. For convenience and simplicity, in the following we hide the subscriptes $j$. 
We thereby adopt a vanilla pre-trained transformer encoder \cite{transformer} to generate representations of claim and candidate evidences. 
Formally, a Transformer-Encoder is fed with a claim or a candidate,
\begin{align}
    \vh_c = \pool(\transformerenc(c; \theta^{(ec)})),\\
    \vh_{x} = \pool(\transformerenc(x; \theta^{(ec)})),
\end{align}
where $\pool(\cdot)$, defined in \cite{bert_devlin_naacl19}, collects the resulting of [CLS] to denote a contextualized representation for the sequence, $\theta^{(ec)}$ denotes the learnable parameters of transformer encoder.

\subsubsection{Competing Veracity Scoring}
In order to detect salient evidences for both parties, we propose to assign two veracity scores for each candidate, namely the ``false score'' and the ``true score'', which represent the level of support for the claim being false or true, respectively. 
Naturally, the two scores can be used for ranking to extract the top-$k$ ``false evidences'' and ``true evidences'' in inference.

To gain the veracity scores, we first calculate a pair of complementary scores for \textit{true} and \textit{false} for each candidate. 
Borrowing common practices from the natural language inference (NLI) area, we apply an interactive concatenation \cite{NLI_Bowman_EMNLP15,sentence-bert_Reimers_EMNLP19} to the pair of claim and candidate, and then perform a neural binary classifier. Formally, we adopt the interactive concatenation written as
\begin{align}
    \vu = [\vh_c; \vh_c \times \vh_{x}; \vh_c - \vh_{x}; \vh_{x}],
\end{align}
where $\vu$ is used to represent the semantic relationship between the claim and candidate. 
Then, a two-way classifier is applied to $\vu$ and produces a two-dimensional categorical distribution corresponding to the \textit{false} and \textit{true} veracity probabilities respectively, i.e.,
\begin{align}
    \vp &= P(z^{(vp)}|\vu; \theta^{(vp)}) \triangleq \softmax(\mlp(\vu; \theta^{(vp)})) , 
\end{align}
where $\mlp(\cdot)$ stands for a multi-layer perceptron, and $\theta^{(vp)}$ is its learnable parameters. 
In this way, the two dimensions of $\vp$
can serve as \textit{false} score $ s^-$and \textit{true} score $s^+$ of the candidate to perform false and true candidate ranking, respectively. 

Nevertheless, merely employing the complementary scores of each candidate for the ranking purpose is suboptimal, as not every sentence carries valuable information or contributes significantly to the final veracity prediction \cite{GenFE_ACL20,zhiwei22coling}. 
To address this limitation, we calculate each candidate’s attention weights for the claim under two competing veracities, enabling a more precise ranking of the evidences.
Formally, we present a concatenation-based attention weight calculation module to gain the false (true) attention weight, 
\begin{align}
    \alpha^v = \frac{\exp(\mlp([\vh_c; \vh_{x}]; \theta^{(v)}))}{\sum_k \exp(\mlp([\vh_c; \vh_{x_k}]; \theta^{(v)})},
\end{align}
where $v$ alternates between $-$ and $+$, $\alpha^-$ and $\alpha^+$ denote false and true weight respectively, $\theta^{(-)}$ and $\theta^{(+)}$ are the learnable parameters. 
As a result, the final \textit{false} (\textit{true}) score of candidate evidence used for ranking is
\begin{align}
    \tilde s^v = \alpha^v s^v.
    \label{eq:sent_scores}
\end{align}

Based on the competing veracity scores of candidate evidences, we can naturally obtain the competing veracity scores of a claim by
\begin{align}
    s_c^v = \sum \tilde s^v, \label{eq:claim_scores}
\end{align}
which can be used to judge the veracity of a claim.

\subsubsection{Training and Inference}
Since the only available supervised data is the veracity label of the claim, we employ two loss objectives to conduct the temporary veracity prediction, taking into consideration the extraction of competing evidences. 

\paragraph{Training. }
The primary objective of this module is to rank and extract two competing sets of evidence using the competing veracity scores. To achieve this, we use a soft objective that considers the competing scores, utilizing the Kullback–Leibler (KL) divergence, 
\begin{gather}
    \gL^{(kl)} = \kl(\vp|| \vp_c), \label{eq:kl_loss} 
\end{gather}
where $\vp_c = (s_c^-, s_c^+)$ is derived from Eq.(\ref{eq:claim_scores}), $\vp$ denotes the pre-defined distribution based on the temporary veracity label $y^{t}$, when $y^{t}$ is \textit{false}, $\vp = (1, 0)$, when $y^{t}$ is \textit{half}, $\vp = (0.5, 0.5)$, and when $y^{t}$ is \textit{true}, $\vp = (0, 1)$. However, this soft KL loss is not suitable for the inference purpose and its connection with the claim is not strong enough. 
Hence, we further define a neural classifier for the temporary three-categorical veracity prediction as
\begin{align}
\begin{split}
     \vp^{(cls)} &= P(z^{(cls)}|\vh_c; \vp_c; \theta^{(cls)})  \\ &\triangleq \softmax(\mlp([\vh_c; \vp_c]; \theta^{(cls)})).
\end{split}
\end{align}
Next, the classification objective to train this module is
\begin{align}
    \gL^{(cls)} = - \sum\nolimits_\gD \log \vp^{(cls)}_{[\hat y^{t} = y^{t}]}, \label{eq:cls_loss}
\end{align}
where $\vp^{(cls)}_{[\hat y^{t} = y^{t}]}$ denotes fetching the probability value corresponding to the temporary veracity label $y^{t}$.

We train the learnable parameters in our proposed extraction module towards a linear combination of the two losses, i.e.,
\begin{align}
    \gL^{(ext)} = \gamma \gL^{(cls)} + (1 - \gamma) \gL^{(kl)}, 
    \label{eq:extractor_loss}
\end{align}
where $\gamma$ is the trade-off weight. 

\paragraph{Inference. }After optimizing the extraction module w.r.t $\gL^{(ext)}$, $\tilde s^-$ and $\tilde s^+$ from Eq.(\ref{eq:sent_scores}) produced in inference can be used as ranking basis. 
For all candidate evidences, they will be ranked twice based on $\tilde s^-$ and $\tilde s^+$ respectively. 
The sets of top-$k$ false and true evidences, i.e., $\gE^-$ and $\gE^+$, are then prepared for the prompt-based reasoning module described in the subsequent section. 

\subsection{Prompt-based Reasoning with LLM} \label{subsec:generator}
To effectively leverage the divergence contained in massive competing evidences on informativeness and soundness, we develop a prompt-based module for a further filter and summarization. 
Motivated by the remarkable abilities of LLMs in reasoning \cite{llama2,GPT-4}, we engage an LLM to conduct abductive reasoning to explain why the claim is false or true based on the extracted evidence sets and a given prior veracity label.

Given a claim $c$, a prior label $\tilde y^v$, and an evidences set $\gE^v$, to prompt the large language model in uniform language modality, we curate a template $T$ that consists of a triplet $\{{c, \tilde y^v}, \gE^v\}$. 
We prompt the LLM with it to generate an explanation $e^v$ that elicits the reasoning knowledge about how to infer the veracity label $\tilde y^v$ based on the interplay of the claim $c$ and the veracity-oriented evidence $\gE^v$. 
Specifically, we design $T$ as:

``
\textit{Given a claim: [$c$], a veracity label [$\tilde y^v$], \
please give me a streamlined rationale associated with the claim, for how it is reasoned as [$\tilde y^v$]. Below are some sentences that may be helpful for the reasoning, but they are mixed with noise: [$\gE^v$]. }
''

The reasoning is performed for both \textit{false} and \textit{true} and thus two reasoning texts are obtained. 
As detailed previously, the evidence set which is consistent with the actual veracity of the claim brings more information and is more reasonable than the competing one. Thus, the LLM prefers to generate solid reasoning in favor of it, while providing weak reasoning with inaccurate information for its competitor, as demonstrated in Appendix \ref{app:validation}. 
In this manner, the two LLM-generated veracity-oriented reasoning $e^-$ and $e^+$, which can be viewed as two explanations to clarify its relevant veracity label, will always possess a relative strength in confidence, greatly facilitating the detection of fake news.

As detailed in \S \ref{sec:related_LLM}, LLMs can provide desirable multi-perspective rationales but still underperform the basic fine-tuned small language models \cite{LLM_SLM_distill_Hu_arxiv23}. Therefore, based on the generated explanations, we further propose a defense-based inference module with an SLM.

\subsection{Defense-based Inference} \label{subsec:predictor}
With the news claim $c$ and two veracity-oriented explanations derived from LLM, we develop a defense-based fake news detector. 
This detector aims to discern the relative strength of the two explanations from their defense, ultimately providing the veracity verdict. 
Concretely, we concatenate the three texts and feed them into a pre-trained Transformer encoder for a contextual representation. Formally, 
\begin{align}
    \vv = &\transformerenc ([c;\text{[SEP]}; e^-; \text{[SEP]}; e^+]; \theta^{(ed)}), 
\end{align}
where [SEP] denotes the special separate token defined in \cite{bert_devlin_naacl19}. Due to the stacked transformer encoders, this representation can effectively capture the semantic differences and connections between these three texts. 
Then, we define a classifier on the top of their rich representation for veracity prediction, 
\begin{align}
\begin{split}
    \vp^{(ver)} &= P(z^{(ver)}|\vv; \theta^{(ver)}) \triangleq \softmax (\mlp (\vv; \theta^{(ver)})).
\end{split}
\end{align}

The training objective of the detection task is written as:
\begin{align}
    \gL^{(ver)} = - \sum\nolimits_\gD \log \vp^{(ver)}_{[\hat y = y]}, \label{eq:veracity_loss}
\end{align}
where $\vp^{(ver)}_{[\hat y = y]}$ denotes fetching the probability value corresponding to the veracity label $y$. 

The inference procedure can be simply written as
\begin{align}
    y^* = \arg \max \vp^{(ver)}.
\end{align}

Based on the model's predictions, we select the corresponding explanation as the final explanation $e$, which can be written as:
\begin{gather}
    e =\left\{
        \begin{array}{lcl}
        e^-,     &      & y^* = \text{false}\\
        e^+,     &      & y^* = \text{true}\\
        e^-; e^+,     &      & y^* = \text{half}
        \end{array} \right. 
        \label{eq:get_explanation}
\end{gather}
Especially, in the case of a ``half'' prediction, to help users understand the false aspect and the true aspect of the claim, we utilize a template to concatenate the two explanations and present both explanations simultaneously. This approach ensures that the final explanation aligns with the predicted veracity label. 

\subsection{Discussion about Generalization} \label{subsec:discuss_gene}
Although our proposed model extracts competing evidences for claims used in reasoning and inference, since its specially designed three-stage architecture, it can be well-generalized to any case, regardless of the presence or absence of competing evidences. Please refer to Appendix~\ref{app:discuss_gene} for more detailed discussions.

\section{Experiment} \label{sec:exp}

In this section, we evaluate L-Defense on two real-world explainable fake news detection benchmarks, and verify the model's effectiveness (\S \ref{subsec:eval_veracity}) and explainability (\S \ref{subsec:eval_explanation}). 
Then, we conduct an extensive ablation study in \S \ref{subsec:ablation} to verify the significance of each proposed module. 
Lastly, in \S \ref{subsec:case_study}, we do case study to show how our proposed model brings improvement.

\paragraph{Datasets. }
We assessed the proposed approach on two explainable datasets, i.e., \textit{RAWFC} and \textit{LIAR-RAW} \cite{zhiwei22coling}, whose statistics are listed in Table \ref{tb:datasets}. \textit{RAWFC} contains the claims collected from Snopes\footnote{\url{https://www.snopes.com/}} and relevant raw reports by retrieving claim keywords. For \textit{LIAR-RAW}, it is extended from the public dataset LIAR-PLUS \cite{LIAR-PLUS_Alhindi_18} with relevant raw reports, containing fine-grained claims from Politifact\footnote{\url{https://www.politifact.com/}}. 
Note that we do not use any debunked justifications in the datasets for both training and inference. 

\paragraph{Training Setups. }
We initialize the encoder in the first module (\S \ref{subsec:extractor}) with RoBERTa$_\text{base}$ \cite{RoBERTa} for the temporary veracity prediction, and set $k = 10$ to extract evidences. 
As for the LLM used in \S \ref{subsec:generator}, we alternate between ChatGPT \cite{chatgpt} and LLaMA2$_{\text{7b}}$ \cite{llama2}. The former refers to a widely used LLM developed by OpenAI, specifically utilizing the ``gpt-3.5-turbo-0613'' version, and the latter is a smaller yet powerful LLM created by Meta AI. 
To obtain the final prediction, we initialize the transformer encoder in \S \ref{subsec:predictor} with the RoBERTa$_\text{large}$. 
Please refer to Appendix~\ref{app:train_setup} for more training details\footnote{The source code is available at \url{https://github.com/wangbo9719/L-Defense_EFND}}.

\begin{table}[t] \small
\caption{\small Summary statistics of datasets. The numbers range from 0 to 5, representing the increasing veracity labels \{pants-fire, false, barely-true, half-true, mostly-true, true\}. ``ALL'' means the total number, and $|\gS|_{avg}$ denotes the average number of sentences associated with each claim. 
} 
\setlength{\tabcolsep}{2pt}
\begin{tabular}{ccccccccccc}
\hline
                          &       & \textbf{0} & \textbf{1} & \textbf{2} & \textbf{3} & \textbf{4} & \textbf{5}  & \textbf{ALL}  & \textbf{$|\gS|_{avg}$}  \\ \hline
{\multirow{3}{*}{\begin{tabular}[c]{@{}c@{}}LIAR-\\ RAW\end{tabular}}}    & train & 812  & 1,985  & 1,611  & 2,087  & 1,950  & 1,647  & 10,065 & 62               \\
                          & eval  & 115  & 259  & 236  & 244  & 251  & 169  & 1,274 & 80                 \\
                          & test  & 86  & 249  & 210  & 263  & 238  & 205 &1,251  & 96                  \\ \hline
\multirow{3}{*}{RAWFC} & train & -  & 514  & -  & 537  & -  & 561  & 1,612 & 154                  \\
                          & eval  & -  & 66  & -  & 67  & -  & 67  & 200 & 156                   \\
                          & test  & -  & 66  & -  & 67  & -  & 67  & 200 & 157                    \\ \hline
\end{tabular}
\label{tb:datasets}
\end{table}

\begin{table}[t] 
\caption{\small Veracity prediction results on RAWFC and LIAR-RAW. $\dagger$Resulting numbers are reported by \citet{zhiwei22coling}, and the results of FactLLaMA are taken from the original paper. The bold numbers denote the best results in each fine-grained genre while the underlined ones are state-of-the performance. }
    \setlength{\tabcolsep}{2pt}
	\centering
	\begin{tabular}{l|ccc|ccc}  \hline
	        & \multicolumn{3}{c|}{\textbf{RAWFC}}           & \multicolumn{3}{c}{\textbf{LIAR-RAW}}    \\ \cline{2-4} \cline{5-7} 
	            & \textbf{P}   & \textbf{R}   & \textbf{macF1}     & \textbf{P}   & \textbf{R}   & \textbf{macF1}  \\ \hline \hline\hline
             
		\multicolumn{7}{l}{\textit{Traditional approach}}\\ \hline
            dEFEND \cite{dEFEND_KDD19}$\dagger$    & 44.93   & 43.26   & 44.07          & 23.09   & 18.56   & 17.51 \\
            SBERT-FC \cite{EXTABS_EMNLP20_Kotonya}$\dagger$  & 51.06 & 45.92 & 45.51    & 24.09   & 22.07   & 22.19 \\
            GenFE \cite{GenFE_ACL20}$\dagger$      & 44.29   & 44.74   & 44.43          & 28.01   & 26.16   & 26.49 \\
            GenFE-MT \cite{GenFE_ACL20}$\dagger$   & 45.64   & 45.27   & 45.08          & 18.55   & 19.90   & 15.15 \\
    		CofCED \cite{zhiwei22coling}$\dagger$  & \textbf{52.99}   & \textbf{50.99}   & \textbf{51.07}          & \textbf{29.48}   & \textbf{29.55}   & \textbf{28.93} \\  
		
		\hline  \hline \hline
  
		\multicolumn{7}{l}{\textit{LLM-based approach}}\\ \hline
		LLaMA2$_{\text{claim}}$         & 37.30   & 38.03   & 36.77         & 17.11    & 17.37   & 15.14       \\
        ChatGPT$_{\text{full}}$         & 39.48   & 45.07   & 39.31         & \textbf{29.64}    & 23.57   & 21.90  \\
        ChatGPT$_{\text{claim}}$       & \textbf{47.72}   & \textbf{48.62}   & \textbf{44.43}         & 25.41    & \textbf{27.33}   & \textbf{25.11}       \\  \hdashline[1pt/1pt]
          
        FactLLaMA   \cite{factllama}      & 53.76   & 54.00   & 53.76         & 32.32    & 31.57    & 29.98 \\
        FactLLaMA$_{\text{know}}$  \cite{factllama}      & \textbf{56.11}   & \textbf{55.50}   & \textbf{55.65}         & \underline{\textbf{32.46}}    & \textbf{32.05}    & \textbf{30.44}       \\ \hline \hline \hline
        \multicolumn{7}{l}{\textit{Ours}}\\ \hline
        L-Denfense$_{\text{LLaMA2}}$    & 60.95     & 60.00     & 60.12     & \textbf{31.63}     & 31.71     & \underline{\textbf{31.40}} \\
        L-Denfense$_{\text{ChatGPT}}$   & \underline{\textbf{61.72}}     & \underline{\textbf{61.01}}    & \underline{\textbf{61.20}}     & 30.55     & \underline{\textbf{32.20}}     & 30.53 \\
		 \hline
	\end{tabular}
	\label{tab:veracity_results}
\end{table}

\subsection{Evaluations on Veracity Prediction} \label{subsec:eval_veracity}

\paragraph{Baselines. }
We compare our L-Defense with two categories, traditional non-LLM-based approaches and LLM-based approaches. 
\textit{Traditional category} contains: 
1) \textbf{dEFEND} \cite{dEFEND_KDD19}; 
2) \textbf{SBERT-FC} \cite{EXTABS_EMNLP20_Kotonya}; 
3) \textbf{GenFE} and \textbf{GenFE-MT} \cite{GenFE_ACL20}; 
4) \textbf{CofCED} \cite{zhiwei22coling}. 
And \textit{LLM-based category} contains: 
5) \textbf{LLaMA2$_{\text{claim}}$} (7b version) \cite{llama2} prompts with the news claim to directly generate a veracity prediction and corresponding explanation; 
6) \textbf{ChatGPT$_{\text{claim}}$} \cite{chatgpt}, which is similar to LLaMA2$_{\text{claim}}$; 
7) \textbf{ChatGPT$_{\text{full}}$} \cite{chatgpt} prompts with the claim and all related reports, and the absence of LLaMA2$_{\text{full}}$ is that the 7b model struggles to produce consistent output after processing such lengthy inputs; 
8) \textbf{FactLLaMA} \cite{factllama} leverages the LORA tuning \cite{lora_ICLR22} to supervised fine-tunes a LLaMA$_{\text{7b}}$ with the claims; 
9) \textbf{FactLLaMA$_{\text{know}}$}, compared with FactLLaMA, fed with external relevant evidence retrieved from search engines.

The veracity prediction results of competitive approaches and ours on the two benchmarks are shown in Table \ref{tab:veracity_results}. 
Following prior works \cite{zhiwei22coling}, we adopt macro-averaged precision (P), recall (R), and F1 score (macF1) to evaluate the performance. 
It is observed that our proposed L-Defense is able to achieve state-of-the-art or competitive performance on the two datasets.

For the traditional approaches, most of them underperform ChatGPT$_{\text{claim}}$, demonstrating the potential of LLMs in fake news detection. 
For the first three LLM-based approaches without any tuning, ChatGPT$_{\text{claim}}$ achieves the best results. LLaMA2$_{\text{claim}}$ loses as its model size is significantly smaller than that of ChatGPT. And a possible reason why ChatGPT$_{\text{full}}$ loses is that the LLM is easily biased by the massive input reports. 
Despite the good performance of ChatGPT$_{\text{claim}}$, it falls short when compared to the best method in the traditional approach, namely CofCED. By contrast, the fine-tuned LLM-based model, i.e., FactLLaMA and FactLLaMA$_{\text{know}}$ achieves the best results in addition to our proposed model. 
This indicates that simply utilizing LLMs for inference yields limited results, while carefully considering how to further leverage LLMs can lead to improved performance. 

Our proposed model makes use of LLM as a reasoner in a novel defense-based framework, and then achieves excellent results on veracity prediction. The improvement is especially significant on RAWFC. Compared with the FactLLaMA$_{\text{know}}$, both versions of our model achieved at least a 4\% enhancement across all metrics. And on LIAR-RAW, although our model achieves inferior results on precision, both variants consistently outperform in terms of macF1. 
Furthermore, in comparison with ours LLaMA2 variant, ours ChatGPT variant achieves slightly superior results on RAWFC while maintaining competitiveness on LIAR-RAW, which suggests that our model does not prioritize the size of the LLM component. A possible reason of this discrepancy is that the $|\gS|_{avg}$ of the two datasets differs significantly, resulting in RAWFC being more information-rich than LIAR-RAW. Consequently, with a much larger scale than LLaMA2, ChatGPT adds more additional information for reasoning, which aligns with the average lengths of explanations as listed in Table \ref{tab:explanation_length} in Appendix \ref{app:explanation_length}.

\subsection{Evaluations on Explanation} \label{subsec:eval_explanation}

\paragraph{Evaluation Metrics. } For the evaluation of explanations, traditional automated evaluation metrics are inadequate to assess the output results of LLMs \cite{survey_LLM_eval}. Fortunately, \citet{llm-for-eval_Chen_arxiv23} demonstrates that ChatGPT excels in assessing text quality from multiple angles, even in the absence of reference texts. Also, some works reveal that ChatGPT evaluation produces results similar to expert human evaluation \cite{LLM-eval_ACL23_Chiang,exp-eval-figure_WWW23_Huang}. 
Therefore, we engage ChatGPT to evaluate the quality of explanations based on four metrics which have been widely employed in human evaluation \cite{zhiwei22coling,exp-eval-metric_IJCAI23_Wang}: \textit{misleadingness}, \textit{informativeness}, \textit{soundness}, and \textit{readability}. A 5-point Likert scale was employed, where 1 represented the poorest and 5 the best in addition to misleadingness. The definitions of the metrics are: 
\begin{itemize}[leftmargin=10pt]
    \item \textbf{Misleadingness (M)} assesses whether the model's explanation is consistent with the real veracity label of a claim, with a rating scale ranging from 1 (not misleading) to 5 (very misleading);
    \item \textbf{Informativeness (I)} assesses whether the explanation provides new information, such as explaining the background and additional context, with a rating scale ranging from 1 (not informative) to 5 (very informative); 
    \item \textbf{Soundness (S)} describes whether the explanation seems valid and logical, with a rating scale ranging from 1 (not sound) to 5 (very sound);
    \item \textbf{Readability (R)} evaluates whether the explanation follows proper grammar and structural rules, and whether the sentences in the explanation fit together and are easy to follow, with a rating scale ranging from 1 (poor) to 5 (excellent). 
\end{itemize}
In order to verify the effectiveness of LLM evaluation, we further propose an automated metric called \textbf{Discrepancy (D)}, which is an objective version of misleadingness and does not consider the quality of the explanation. 
It is obtained by calculating the absolute difference between the predicted and actual labels' scores. Specifically, for RAWFC, the scores for the three labels are [0, 2.5, 5]; for LIAR-RAW, the scores for the six labels are [0, 1, 2, 3, 4, 5]. The larger the discrepancy between the predicted and true labels, the higher the score, indicating a greater degree of misleading.

\paragraph{Baselines. } 
Observed on the veracity prediction results in Table \ref{tab:veracity_results}, we propose the following baselines: 
1) \textbf{Oracle} generates an explanation for why the claim is classified as its actual veracity label, by providing both the claim and the actual veracity label to ChatGPT; 
2) \textbf{CofCED} \cite{zhiwei22coling}, the best model in traditional approaches; 
3) \textbf{ChatGPT$_{\text{claim}}$}, which performs better than the LLaMA2 version; 
4) \textbf{ChatGPT$_{\text{full}}$}. 
To make a fair comparison, we limited the number of extracted sentences for CofCED. The length of explanations generated by each model can be found in the Appendix \ref{app:explanation_length}. 
To gain the explanations of our proposed model on LIAR-RAW, we categorize \textit{pants-fire}, \textit{false}, and \textit{barely-true} as \textit{false}, \textit{half-true} remains as \textit{half-true}, and \textit{mostly-true} and \textit{true} are viewed as true. As a result, the explanation of each veracity can be derived based on Eq.(\ref{eq:get_explanation}).

The evaluation results for the quality of explanations are presented in Table \ref{tab:gpt_explanation_eval}, showing that L-Defense consistently achieves excellent performance. 
The alignment of score trends on \textit{M} and \textit{D} provides partial validation of the effectiveness of our evaluation methodology. 
The Oracle prompted with the actual veracity achieves superiority across almost all metrics, which can be viewed as a ceiling of explanation quality. 
Compared to it, L-Defense achieves superior or comparable performance, highlighting the excellence of our method in generating human-read explanations. 

Since the CofCED generates explanations by extracting from reports, the discrete evidence sentences are hard to fit together and may overlap. Therefore, its explanation achieves the worst performance compared with other LLM-based models. Also, its incoherent explanation leads to a high misleadingness score, though the discrepancy score is much better. 
This demonstrates the significance of coherence in generating a streamlined and understandable explanation. 
In terms of two versions of ChatGPT, ChatGPT$_{\text{full}}$ exhibits better performance than ChatGPT$_{\text{claim}}$ on RAWFC for the last three metrics. However, it underperforms ChatGPT$_{\text{claim}}$ on LIAR-RAW across all metrics. 
One possible reason is that in the full version, the input length for RAWFC is approximately twice that of LIAR-RAW, thereby incorporating more information. At the same time, this also indicates that ChatGPT is highly vulnerable to the input context. The introduction of additional raw reports, in contrast to ChatGPT$_{\text{claim}}$, leads to a completely different performance in ChatGPT$_{\text{full}}$. 
For the two variants of our proposed L-Defense, the ChatGPT variant always beats the LLaMA2 variant on misleading-related metrics while losing in the latter three metrics. 
It further demonstrates that the performance of our model is not limited by the size of the LLM. 

\begin{table}[t] \small
\setlength{\tabcolsep}{1.2pt}
\caption{\small Evaluation results of explanation quality using a 5-Point Likert scale rating by ChatGPT on RAWFC and LIAR-RAW. For metrics D and M, a lower score indicates better performance, while a higher score indicates better performance for the remaining metrics. The bold numbers denote the best results in addition to Oracle while the underlined ones are better than Oracle. } 
\centering
    \begin{tabular}{l|c|cccc|c|cccc}
    \hline
                      & \multicolumn{5}{c|}{\textbf{RAWFC}}          & \multicolumn{5}{c}{\textbf{LIAR-RAW}}     \\ \cline{2-11} 
                      & \textbf{D} & \textbf{M}     & \textbf{I}     & \textbf{S}     & \textbf{R}                & \textbf{D}  & \textbf{M}    & \textbf{I}    & \textbf{S}    & \textbf{R}    \\ \hline
    Oracle            &  -            & 1.52  & 4.46  & 4.73   & 4.72      & -      & 1.85 & 4.44 & 4.60  & 4.69 \\ \hline
    CofCED \cite{zhiwei22coling}   & 1.53        & 2.74    & 2.89          & 1.93   & 2.46    & 1.33 & 3.64 & 1.75 & 1.76  & 1.59 \\
    ChatGPT$_\text{full}$          & 1.81  & 2.07  & \underline{\textbf{4.44}} & 4.62  &\textbf{ 4.69}        & 1.39   & 2.29 & 3.71 & 4.04  & 3.99 \\
    ChatGPT$_\text{claim}$         & 1.70 & 1.97   & 4.00  & 4.44   & 4.68       & 1.39   & 2.27 & 3.93 & 4.29  & 4.50 \\ \hline
    L-Defense$_\text{LLaMA2}$ & \textbf{1.30}  & 1.95 & \underline{\textbf{4.44}} & \textbf{4.67}   & 4.62         & 1.36   & 2.20 & \textbf{4.39} & \underline{\textbf{4.64}}  & \textbf{4.63} \\
    L-Defense$_\text{ChatGPT}$  & \textbf{1.30}  & \textbf{1.91}  & 4.17  & 4.41   & 4.49     & \textbf{1.31}   & \textbf{2.06} & 4.12 & 4.28  & 4.47 \\ \hline
    \end{tabular}
    \label{tab:gpt_explanation_eval}
\end{table}

\begin{table}[t] \small
\setlength{\tabcolsep}{3pt}
\caption{\small Explanation evaluation results using a 5-Point Likert scale rating by both ten human annotators and ChatGPT on 30 randomly sampled samples from RAWFC's testset. Scores from ten annotators were averaged. The bold numbers denote the best results in addition to Oracle.} 
\centering
    \begin{tabular}{l|cccc|cccc}
    \hline
                      & \multicolumn{4}{c|}{\textbf{ChatGPT}}          & \multicolumn{4}{c}{\textbf{Human}}     \\ \cline{2-9} 
                       & M     & I     & S     & R                     & M    & I    & S    & R    \\ \hline
    Oracle                        & 1.53 & 4.50 & 4.77 & 4.77    & 1.47 & 3.61 & 3.89 & 3.86 \\ \hline
    CofCED \cite{zhiwei22coling}   & 2.90 & 2.77 & 2.87 & 2.47  & 2.46  & 2.91 & 2.47 & 2.44  \\
    ChatGPT$_\text{full}$     & 2.07    & 4.43 & \textbf{4.67} & \textbf{4.73}    & 2.22  & 3.22 & 3.38 & \textbf{3.57}  \\
    ChatGPT$_\text{claim}$    & 2.33    & 4.17 & 4.43  & 4.63   & 2.68  & 2.68 & 2.84 & 3.27  \\ \hline
    L-Defense$_\text{LLaMA2}$ & 1.87 & \textbf{4.50} & \textbf{4.67} & 4.67       & 2.12 & 3.48 & 3.37 & 3.49  \\
    L-Defense$_\text{ChatGPT}$  & \textbf{1.77}  & 4.40 & 4.60 & 4.53    & \textbf{1.97}  & \textbf{3.68} & \textbf{3.52} & 3.56  \\ \hline
    \end{tabular}
    \label{tab:human_explanation_eval}
\end{table}

\begin{figure}[t]
    \centering
    \includegraphics[width=0.4\textwidth]{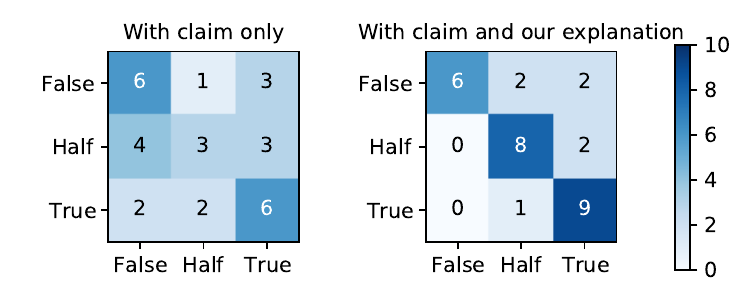}
    \vspace{-4mm}
	\caption{\small Confusion matrixes of the judgment results made by 10 annotators on 30 randomly sampled samples. The left one is derived by providing annotators with only the claim, while the right one is derived by providing annotators with both the claim and the explanations generated by our L-Defense. The results from 10 annotators were averaged and rounded off.}
	\label{fig:human_eval} 
\end{figure}

\paragraph{Human Evaluation. }

To further validate the effectiveness of the LLMs evaluation and the helpfulness of our proposed model, we asked ten English-speaking adult volunteers as annotators and thenconducted two human evaluations for the explanation quality. 
On the one hand, we instructed the annotators to perform similar evaluations as ChatGPT did. 
As shown in Table~\ref{tab:human_explanation_eval}, the evaluative results of the annotators on various metrics are largely consistent with the ranks made by ChatGPT. The main difference is that humans prefer the explanations made by the ChatGPT variant of our model while ChatGPT prefers the LLaMA2 variant. 
On the other hand, as shown in Figure \ref{fig:human_eval}, our proposed model remarkably reduces error judgments and aids humans in understanding truth. 

\begin{table}[] \small
\caption{\small Ablation study of veracity prediction on RAWFC. }
    \centering
    \begin{tabular}{l|ccc}
\hline
\textbf{Method}                & \textbf{P} & \textbf{R} & \textbf{F1} \\ \hline 
L-Denfense$_{\text{LLaMA2}}$    & 60.95     & 60.00     & 60.12       \\ \hline

~~w/o evidences         & 54.45  & 52.56  & 52.51   \\
~~with random evidences      & 57.09  & 56.47  & 56.35   \\ 
~~w/o prior label       & 55.97  & 56.02  & 55.98   \\
~~w/o explanations               & 52.92  & 51.96  & 51.83    \\ 
~~w/o inference training & 39.30  & 38.88  & 29.71   \\ \hline
\end{tabular}
    
    \label{tab:ablation}
\end{table}

\begin{table*}[t] \small
\centering
    \caption{\small Case study. The claim is selected from RAWFC testset. Four methods' predictions and explanations are listed. The ``Gold'' method denotes the judgement from Snopes. We display the top-3 evidences extracted by our extraction module and CofCED. The fragments supporting the claim as true are in green background, while the fragments supporting the claim as false are in red background.}
    \begin{tabular}{p{17.5cm}}
    \hline
    \multicolumn{1}{l}{\makecell[l]{\textbf{\href{https://www.snopes.com/fact-check/obama-coronavirus-masks/}{Claim}}: Former President Barack Obama's administration was to blame for the shortage of protective equipment like N95 respirator masks in the early \\months of the 2020 COVID-19 pandemic.}
} \\ \hline

        \textbf{Method}: Gold, \textbf{Prediction}: \color{orange}{half}  \\ \hdashline
        
        $\cdot$ What's true: The U.S. federal stockpile of N95 protective face masks was \sethlcolor{mylightgreen}\hl{largely depleted during the 2009 swine flu outbreak and was not restocked}.   \\ 
        $\cdot$ What's false: However, the Obama administration \sethlcolor{pink}\hl{was not solely responsible for the current shortage of masks}. In the intervening years, the stockpile went unreplenished as \sethlcolor{pink}\hl{the Trump administration failed to heed indications that dramatic shortages could occur}. \\ \hline

        \textbf{Method}: L-Defense$_{\text{ChatGPT}}$, \textbf{Prediction}: \color{orange}{half}  \\ \hdashline
        
         $\cdot$ Extracted true-supporting evidences: (1) WASHINGTON — Senior Trump administration official on Thursday blame a shortfall of N95 mask early in the coronavirus pandemic on \sethlcolor{mylightgreen}\hl{mismanagement after the 2009 swine flu pandemic under former President Barack Obama}. (2) Health care professional and expert say we now have \sethlcolor{mylightgreen}\hl{a shortage of mask}, gown, glove, ... . (3) ..., \sethlcolor{mylightgreen}\hl{a scant 1 percent} of the estimate 3.5 billion mask the nation would need in a severe pandemic. \\ 
       $\cdot$ Extracted false-supporting evidences:  (1) \sethlcolor{pink}\hl{Mr. Trump appear intent on focus attention on the Obama administration} at a time when his own handling of the outbreak have come under intense criticism. (2) And \sethlcolor{pink}\hl{Trump attempt to blame a president who leave office more than three year ago for his failure to act two month ago} might be his most pathetic blame-shirking yet. (3) Dr. Luciana Borio, ...   \\  \hdashline
       
       $\cdot$ What's true: \sethlcolor{mylightgreen}\hl{A senior Trump administration official attributed the shortfall of N95 masks to mismanagement following the 2009 swine flu pandemic under Obama}. ...  \sethlcolor{mylightgreen}\hl{the stockpile of masks was significantly below the estimated amount needed for a severe pandemic}. ...  \\

       $\cdot$ What's false: The sentences provided contain statements that \sethlcolor{pink}\hl{criticize President Trump for attempting to shift blame onto the Obama administration}, but they do not provide any concrete evidence or specific actions taken by the Obama administration that would have caused the shortage. ...  \\    \hline

       \textbf{Method}: CofCED, \textbf{Prediction}: \color{red}{false}  \\ \hdashline
        (1) The World Health Organization have say mask, goggles and other protective equipment use by health worker be in short supply due to ``rise demand, hoard and misuse.'' (2) "Let's stay calm, listen to the experts, and follow the science." tweets The death toll in the US from the virus rose to 11 on Wednesday. (3) Washington: \sethlcolor{pink}\hl{Former US president Barack Obama} call Wednesday for people to take ``common sense precaution'' over the coronavirus outbreak--\sethlcolor{pink}\hl{advising them to follow hand-washing guideline but not to wear masks}.  \\ \hline

        \textbf{Method}: ChatGPT$_{\text{full}}$, \textbf{Prediction}: \color{red}{false}  \\ \hdashline
        The reports provided primarily discuss \sethlcolor{pink}\hl{the shortage of protective equipment and the lack of preparedness in general, but they do not attribute the blame solely to the Obama administration}. ... . They also highlight \sethlcolor{pink}\hl{the failure of the current administration to replenish the stockpile after the H1N1 outbreak}. ... \\

         \hline
    \end{tabular} \label{tab:case_study}
\end{table*}

\subsection{Ablation Study} \label{subsec:ablation}
To evaluation the contribution of each component, we conduct an extensive ablation study for L-Denfense$_{\text{LLaMA2}}$ on veracity prediction by removing or replacing the key component: 
1) ``w/o evidences'' removes the extracted competing evidences $\gE^v$ from the prompt template \textit{T} for LLM; 
2) ``with random evidence'' replaces the $\gE^v$ with random sampled sentences in \textit{T}; 
3) ``w/o prior label'' removes the given prior veracity label $\tilde y^v$ in \textit{T}; 
4) ``w/o explanations'' replaces the LLM-generated veracity-oriented explanations $e^v$ with corresponding extracted competing evidences $\gE^v$ in defense-based inference (\S \ref{subsec:predictor}); 
5) ``w/o inference training'' replaces the fine-tune process (\S \ref{subsec:predictor}) with ChatGPT's predictions.

As shown in Table \ref{tab:ablation}, it is observed that the original versions significantly outperform their component-deprived versions. 
The degradation of ``w/o evidences'' and ``with random evidences'' demonstrate the necessity of the evidence extraction module. 
Moreover, despite the ``with random evidences'' being inferior to L-Denfense, it still achieved commendable performance. This indicates the superiority of our proposed defense-based framework and the powerful ability of LLM. 
Furthermore, the results of ``w/o prior label'' argue the importance of an enforced prior label to instruct an LLM. 
And the results of ``w/o explanations'' prove the necessity of the prompt-based reasoning module. 
Finally, the significant performance degradation of ``w/o inference training'' reveals a clear gap between LLM and fine-tuned SLM in veracity prediction. Also, it highlights that the final step of our framework is indispensable.

\subsection{Case Study} \label{subsec:case_study}

\textit{Does the evidence extraction module successfully separate two competing parties from the sea of wisdom? }
As shown in Table \ref{tab:case_study}, the top-3 true evidences and false evidences extracted by L-Defense focus on the different viewpoints of the claim. 
The true evidences highlight the mask shortage caused by mismanagement following the 2009 swine flu pandemic under Obama. The false evidences indicate Trump failed to pay attention to the shortage and suggest the news may stem from Trump's influence. These two competing evidence sets are consistent with their respective gold justifications, which verifies the effectiveness of the evidence extraction module.

\textit{Does the two competing justifications generated by the prompt-based module help to determine veracity? }
Based on the extracted competing evidences, the two justifications generated from the LLM reasoning show an obvious competition on the veracity of the claim. Since their strengths in confidence are equivalent in general, the defense-based inference module gives a correct prediction. 

\textit{Does the defense-based strategy mitigate the majority bias problem? }
With the same information provided, only our proposed model makes the correct prediction. The top-ranked evidences from CofCED do not contain any evidence about the true aspect and is biased by the third evidence, causing a misclassification. The ChatGPT$_{\text{full}}$ gives a justification that ignores the true points and predicts the claim as false, due to the lack of deep thinking or training. In contrast, our framework avoids these problems benefiting from our novel defense-based strategy.

\section{Conclusion} \label{sec:conclusion}
In this work, we propose a novel defense-based framework by effectively leveraging the competing wisdom inherent in raw reports for explainable fake news detection. Specifically, we first propose an evidence extraction module to detect salient evidence for two competing parties. 
Since the extracted competing evidence is diverse and massive, we then design a prompt-based module integrating the powerful reasoning ability of LLM to generate streamlined justifications for two possible veracities. Finally, we determine the veracity of a claim by modeling the defense among these justifications and give the final explanations based on the prediction. 
The experiments on two real-world benchmarks can greatly support our motivations, and empirical results show state-of-the-art performance with explainability. 

\section*{Acknowledgement}
The authors would like to thank the anonymous reviewers for their valuable and helpful comments. This work is supported by the National Key Research and Development Program of China (No. 2023YFF0905400), the National Natural Science Foundation of China (No. U2341229, No. 62206233), the Science and Technology Development Program of Jilin Province (No. 20210508060RQ), and Hong Kong RGC ECS (No. 22200722).

\bibliographystyle{ACM-Reference-Format}
\bibliography{ref}

\appendix

\section{The Measuring of Majority Bias}
To measure the majority bias, for each claim, we calculate the ratio of sentences among all candidate sentences that are semantically similar to the extracted evidences, which can be written as: 
$$ r = \frac{1}{k} \sum_{i=1}^{k} (\frac{1}{m} \sum_{j=1}^{m} \mathbb{I}(\text{cos}(\boldsymbol{h}_i, \boldsymbol{h}_j)>0.5)), $$
where $k$ denotes the number of extracted evidences, $m$ denotes the number of all candidate sentences, $\mathbb{I}(\cdot)$ is an indicator function that returns 1 when the condition in $\cdot$ is true and 0 otherwise, $\boldsymbol{h}_i$ and $\boldsymbol{h}_j$ denote the representations of extracted evidence and candidate sentence derived from SBERT \cite{sentence-bert_Reimers_EMNLP19} respectively, $\text{cos}(\cdot)$ denotes the cosine similarity computation, and 0.5 is a threshold. A larger value of this ratio indicates a higher degree of suffering from majority bias. Based on the test set of RAWFC, the average ratios of our main baseline CofCED \cite{zhiwei22coling} and our L-Defense are 0.32 and 0.22 respectively (we will add this result in the paper revision). Moreover, as shown in Table \ref{tab:human_explanation_eval}, the results of human evaluation on the Informativeness (I) also suggest that the evidences extracted by CofCED are much less informative compared with ours, since the CofCED easily falls into a narrow perspective, i.e., the major perspective. To handle this bias, our proposed defense-based approach provides an effective solution. 

\section{Detailed Discussion about Generalization} \label{app:discuss_gene}

As we detailed in \S \ref{sec:approach}, the proposed L-Defense is a three-stage model. For the first stage, the competing evidence extraction module assigns two competing scores for each candidate evidence and then ranks the true-oriented and false-oriented evidence sets respectively. Thus, whether a claim contains competing evidences or not, the two evidence sets can be obtained. If a claim has no competing evidences, there are two cases: (1) no evidences supports the claim as true, the stances of all evidences are mixed with the false-oriented and neutral (i.e., no-true), and (2) no evidences supports the claim as false, the stances of all evidences are mixed with the true-oriented and neutral (i.e., no-false). For the ``no-true'' (or ``no-false'') case, regardless of the performance of the extraction module, both evidence sets contain evidences to support its false (or true) veracity or neutral.
In the second stage, since a prior veracity label is used to prompt the LLM, the LLM will give a rationale according to the label (\S \ref{subsec:generator}). Without true-oriented (or false-oriented) evidences, the LLM would generate inconsistent or untenable rationales for the true-oriented (or false-oriented) party.
In the third stage, based on the obvious differences between the two competing explanations in terms of soundness and informativeness, the well-trained defense-based inference module will easily make an accurate prediction (\S \ref{subsec:predictor}).
Therefore, our model can be applied to the claims with/without competing evidences.

\section{Implementation Details} \label{app:train_setup}
\subsection{Training Setup}
For the two trained modules, we use a mini-batch Stochastic Gradient Descent (SGD) to minimize the loss functions, with Adam optimizer, 10\% warm-up, and a linear decay of the learning rate. Other hyper-parameters used in training are listed in Table \ref{tab:hyper_parameters}. 
For the training of evidence extractor on LIAR-RAW, the temporary veracity labels are assigned by categorizing \textit{pants-fire}, \textit{false}, and \textit{barely-true} as false, \textit{half-true} as
half-true, and \textit{mostly-true} and \textit{true} as true. 
For the prompt-based reasoning, we set the temperature to 0.8, allowing the LLM to flexibly apply the provided evidence and its own knowledge for a rich justification. 
And we set the temperature as 0 during the explanation evaluation. 
Veracity prediction results are the best values from ten runs. 
All experiments were conducted using a single A40 GPU. 

\subsection{Prompt Designing}
For the system prompt to the ChatGPT and LLaMA2 in the proposed prompt-based module, we design the message as:
For the system prompt to the ChatGPT and LLaMA2 in the proposed prompt-based module, we designed it with the fundamental concept of addressing three key aspects: \textit{What are you? What should you do? And what is your goal?} For the user prompt, we integrate all needed elements into the instruction. Following the initial design, we used these prompts to check the format of the LLM's output by testing them on randomly selected samples from the training set. Our criterion for success was to obtain a streamlined rationale that enables readers to assess the claim's veracity without relying on additional background knowledge. Based on the LLM's output, we iteratively refined the prompts through several rounds of revisions until arriving at the current version. The final user prompt we used is listed in the in \S \ref{subsec:generator}. And the final system prompt is:

``\textit{You have been specially designed to perform abductive reasoning for the fake news detection task. Your primary function is that, according to a veracity label about a news claim and some sentences related to the claim, please provide a streamlined rationale, for how it is reasoned as the given veracity label. Note that the related sentences may be helpful for the explanation, but they are mixed with noise. Thus, the rationale you provided may not necessarily need to rely entirely on the sentences above, and there is no need to explicitly mention which sentence was referenced in your explanation. Your goal is to output a streamlined rationale that allows people to determine the veracity of the claim when they read it, without requiring any additional background knowledge. The length of your explanation should be less than 200 words. }''

\section{More Experiments}

\subsection{Validation of Assumption} \label{app:validation}
\textit{Whether the assumption that the party indicating the truth is more informative and sounder than its competitors is true. }
As shown in Table \ref{tab:assumption_eval}, in a quantitative view, the comparable results between the true and false justifications across different classes support our assumption. 
When the claim is false, the evaluation results of false justification are better than those of the true one; when the claim is half-true, both yield competitive results; and when it is true, the true-oriented one is better. 

\subsection{The Length of Explanations} \label{app:explanation_length}
We list the average length of explanations generated by ours and all baselines in Table \ref{tab:explanation_length}. For a fair comparison, we selected the top-6 sentences ranked by CofCED \cite{zhiwei22coling} as its explanation. We limited the generated length of other LLM-based models to 200 words in our designed system prompt as L-Defense. Since the explanation corresponding to the ``half'' prediction of L-Defense is derived from the combining of two competing justifications, its length is slighter longer than others. 

\subsection{Ablation Study of Extraction Objective}
As shown in Table \ref{tab:ablation_objective}, the results demonstrate the contribution of each objective in the extraction module. 

\begin{table}[t] \small
    \centering
    \caption{\small Hyper-parameters. }
    \begin{tabular}{lcc}
\hline
\textbf{Hyperparm}           & \textbf{RAWFC}       & \textbf{LIAR-RAW}      \\ \hline \hline
\multicolumn{3}{l}{Competing Evidence Extraction} \\ \hline 
Epoch               & 5           & 5             \\
Batch Size          & 2           & 2             \\
Learning Rate       & 1e-5        & 1e-5          \\
$\gamma$ in Eq.(\ref{eq:extractor_loss})           & 0.9         & 0.5            \\ \hline
\multicolumn{3}{l}{Defense-based Inference}       \\ \hline
Epoch               & 5           & 5             \\
Batch Size          & 8           & 32            \\
Learning Rate       & 5e-6        & 5e-6        \\ \hline
\end{tabular}
    \label{tab:hyper_parameters}
\end{table}

\begin{table}[t] \small
\caption{\small Explanations evaluation results of competing explanations using a 5-Point Likert scale rating by ChatGPT on RAWFC's test set. }
\begin{tabular}{l|cc|cc|cc}
\hline
\textbf{Gold veracity label}  & \multicolumn{2}{c|}{\textbf{False}} & \multicolumn{2}{c|}{\textbf{Half}} & \multicolumn{2}{c}{\textbf{True}} \\ \hline
Given prior label  & F            & T           & F           & T           & F           & T          \\ \hline
Informativeness & 4.06        & 3.95            & 3.98        & 4.28        & 3.85        & 4.46           \\
Soundness       & 4.21        & 3.88            & 4.09        & 4.10        & 3.92        & 4.45           \\ \hline
\end{tabular} \label{tab:assumption_eval}
\end{table}

\begin{table}[t] \small
\caption{\small The average number of tokens per explanation generated by each method on the RAWFC and LIAR-RAW test sets. }
\begin{tabular}{lcc}
\hline
\textbf{Method}           & \textbf{RAWFC} & \textbf{LIAR-RAW} \\ \hline
Oracle       & 201.68               & 220.75                  \\
CofCED \cite{zhiwei22coling}       & 298.48               & 220.56                  \\
ChatGPT$_{\text{full}}$ & 144.32               & 139.15                  \\
ChatGPT$_{\text{claim}}$      & 128.71               & 150.97                  \\
L-Defense$_{\text{ChatGPT}}$    & 266.61               & 225.52                  \\
L-Defense$_{\text{LLaMA2}}$    & 305.50               & 175.38                  \\ \hline
\end{tabular} \label{tab:explanation_length}
\end{table}

\begin{table}[t] \small
\caption{\small Ablation study on objective detailed in  of veracity prediction on RAWFC. }
    \centering
    \begin{tabular}{l|ccc}
\hline
\textbf{Method}                & \textbf{P} & \textbf{R} & \textbf{F1} \\ \hline 
 L-Defense$_\text{extractor}$ & 51.06  & 50.95  & 50.69    \\ \hline

 ~~w/o KL-divergence obj           & 48.41  & 46.95  & 46.79  \\
 ~~w/o classification obj          & 47.08  & 46.52  & 46.75   \\ \hline 

\end{tabular} \label{tab:ablation_objective}
\end{table}

\begin{table*}[t] \footnotesize
\centering
    \caption{\small More case studies. The ``Gold'' method denotes the gold label and explanations from Snopes. The two explanations generated by L-Defense are all listed for comparison. }
    \begin{tabular}{p{17cm}}
    \hline 
    
        \multicolumn{1}{l}{\makecell[l]{\textbf{\href{https://www.snopes.com/fact-check/supreme-court-pelosi/}{Claim}}: The U.S. Supreme Court has ruled against House Speaker Nancy Pelosi on impeachment.}} \\ \hline

        \textbf{Method}: Gold, \textbf{Prediction}: \color{red}{false} \\ \hdashline
         A blog post that went viral in December 2019 created the false impression that the U.S. Supreme Court had issued a "landmark decision" favoring President Donald Trump in ongoing impeachment proceedings against him, thereby dealing a blow to House Speaker Nancy Pelosi. The post, "Supreme Court Ends Pelosi’s Reign of Terror with Landmark Decision on Impeachment," was published on Dec. 17, 2019, and reported that "Nancy Pelosi just got spanked by the Supreme Court in a new impeachment ruling." However, the story did not support what the headline proclaimed.   \\   \hline
        
        \textbf{Method}: L-Defense$_{\text{ChatGPT}}$, \textbf{Prediction}: \color{red}{false} \\ \hdashline
         $\cdot$ What's true: The claim is reasoned as True based on the following rationale: House Speaker Nancy Pelosi has been vocal in her opposition to President Trump's Supreme Court nominee, Amy Coney Barrett. Pelosi referred to Barrett as an "illegitimate Supreme Court justice" during a news conference. This indicates a disagreement between Pelosi and the Supreme Court, suggesting that the Supreme Court has ruled against Pelosi's position on impeachment. Additionally, Pelosi's statement about not ruling out impeaching President Trump again to delay the Senate filling the Supreme Court vacancy implies that she is concerned about the Court's influence on future impeachment proceedings. Given Pelosi's opposition and concerns, it is reasonable to conclude that the U.S. Supreme Court has ruled against her on impeachment.  \\
         $\cdot$ What's false: The claim that the U.S. Supreme Court has ruled against House Speaker Nancy Pelosi on impeachment is false. There is no evidence or mention of such a ruling in the provided sentences. The sentences do mention various topics related to the Supreme Court, such as Joe Biden's stance on court packing and the court's upcoming hearing on the Trump administration's challenge to Obamacare. However, none of these sentences support the claim that the Supreme Court has ruled against Pelosi on impeachment. Additionally, there is a mention of Pelosi's comment on the Supreme Court's ruling on financial records of the President, but this does not pertain to impeachment. Therefore, based on the lack of any supporting evidence or mention of such a ruling, the claim is deemed false. \\    \hline \hline

       \multicolumn{1}{l}{\makecell[l]{\textbf{\href{https://www.snopes.com/fact-check/lions-kill-poacher-south-africa/}{Claim}}: In February 2018, a pride of lions killed and ate a man suspected of being a poacher near Kruger National Park in South Africa.}} \\ \hline

       \textbf{Method}: Gold, \textbf{Prediction}: \color{green}{true} \\ \hdashline
        Ingwelala Private Nature Reserve in South Africa was shaken with the screams of a man who was attacked and killed by [a] Pride of Lion[s] in [a] private game reserve near the Kruger National Park. Before the lions were dispersed with the help of gunshots, they had eaten most of the body barring the head that was untouched. \\   \hline

        \textbf{Method}: L-Defense$_{\text{ChatGPT}}$, \textbf{Prediction}: \color{green}{true} \\ \hdashline
        $\cdot$ What's true: The presence of sentences discussing illegal hunting and poaching activities in Kruger National Park, as well as the mention of a man being suspected of being a poacher, provide evidence that supports the claim. The sentence mentioning the man's family seeking help from authorities after his death further strengthens the claim. Additionally, the sentence about the man and two others hunting illegally in the park and being surprised by an elephant suggests that the man's death was a result of his illegal activities. The mention of previous incidents where suspected poachers have fallen victim to their prey in the park also supports the claim. Overall, the combination of these sentences provides a rationale for why the claim is reasoned as true. \\
       $\cdot$ What's false: The claim is reasoned as False because there is no evidence to support the claim that a pride of lions killed and ate a man suspected of being a poacher near Kruger National Park in South Africa in February 2018. The sentences provided mention incidents of lion poaching and rhino poaching in the region, but there is no direct mention or evidence of a lion killing a suspected poacher. The sentences also discuss the Game Theft Act and the history of poaching in the area, but they do not provide any information about the specific incident mentioned in the claim. Additionally, there is mention of a lion skeleton being sold and the methods used by poachers, but again, no direct evidence of the claim. Therefore, based on the lack of supporting evidence, the claim is deemed False.
)  \\    \hline \hline

        \multicolumn{1}{l}{\makecell[l]{\textbf{\href{https://www.snopes.com/fact-check/harvard-study-apple-iphones-sales-slows/}{Claim}}: A Harvard study proved that Apple purposely slows down its older model iPhones to coincide with device releases and to boost sales of \\ new models.}} \\ \hline
        
        \textbf{Method}: Gold, \textbf{Prediction}: \color{orange}{half} \\ \hdashline
        $\cdot$ What's true: Some recent iOS upgrades slow down the processing speeds of iPhones with degraded batteries.
        
        $\cdot$ What's false: The processor slowdown is (according to Apple) intended to prevent unexpected power-related shutdowns and other problems in older iPhones that have batteries whose performance has degraded over time, not to boost sales of just-released newer iPhone models. \\   \hline

        \textbf{Method}: L-Defense$_{\text{ChatGPT}}$, \textbf{Prediction}: \color{red}{false} \\ \hdashline
        $\cdot$ What's true: The presence of a study conducted by a Harvard University PhD student lends credibility to the claim that Apple purposely slows down its older model iPhones to coincide with device releases and boost sales of new models. This study supports long-standing suspicions among customers that the company engages in planned obsolescence to encourage upgrades. Additionally, the fact that Apple ships fewer iPhones in recent quarters suggests that the company may employ tactics to incentivize customers to purchase newer models. The negative reaction from consumers towards planned obsolescence, as well as the potential frustration for gamers who do not use Apple products, adds further weight to the claim. The study's affiliation with Harvard University enhances its reliability, as the institution is recognized for its academic rigor. All these factors combined provide a streamlined rationale for the claim being true.  \\
       $\cdot$ What's false: The claim that Apple purposely slows down its older model iPhones to coincide with device releases and boost sales of new models is reasoned as False. A Harvard study is mentioned, but the sentences provided are unrelated or do not support the claim. The sentences mention a PHD student's suspicion, a study on employee performance, the new Apple Watch's health features, frustrations with new iPhone features, a description of the new camera feature, Tim Cook's product announcement, complaints about slow performance after software updates, and a discussion of finding in a New York Times column. None of these sentences explicitly support the claim or provide evidence that Apple intentionally slows down older iPhones. Hence, the rationale for the claim being False is that there is no supporting evidence or relevant information to substantiate it.  \\    \hline

         \hline
    \end{tabular} \label{tab:more_cases}
\end{table*}

\subsection{More Case Studies}
We provide more cases of the L-Defense's predictions and explanations in Table \ref{tab:more_cases}. 
In general, the first two cases show that the informativeness and soundness of the truth side are at a higher level than the competing one, which proves our assumption. Moreover, we analyze error cases made by L-Defense on RAWFC's test set, most of them fall into the failure due to the failure extraction of accurate evidence sentences, as the third case listed in Table \ref{tab:more_cases}. 

\section{Future Work}
The advanced method to extract salient evidence can be further explored to provide accurate evidences and then gain improvements. Also, the advanced prompt method to improve the reasoning ability in fake news detection of LLM is valuable to explore. Furthermore, how to adaptively select suitable LLM for different claims can be investigated for further improvement.

\end{document}

%% file: math_commands.tex
\usepackage{amsmath,amsfonts,bm}

\def\eqref#1{equation~\ref{#1}}

\def\1{\bm{1}}

\def\vh{{\bm{h}}}

\def\vp{{\bm{p}}}

\def\vu{{\bm{u}}}
\def\vv{{\bm{v}}}

\DeclareMathAlphabet{\mathsfit}{\encodingdefault}{\sfdefault}{m}{sl}
\SetMathAlphabet{\mathsfit}{bold}{\encodingdefault}{\sfdefault}{bx}{n}

\def\gD{{\mathcal{D}}}
\def\gE{{\mathcal{E}}}

\def\gL{{\mathcal{L}}}

\def\gS{{\mathcal{S}}}

\def\gX{{\mathcal{X}}}

\newcommand{\softmax}{\mathrm{softmax}}

